\documentclass[10pt,twocolumn,letterpaper]{article}

\usepackage{cvpr}              % To produce the CAMERA-READY version
\usepackage[table]{xcolor}

\definecolor{cvprblue}{rgb}{0.21,0.49,0.74}
\usepackage[pagebackref,breaklinks,colorlinks,citecolor=cvprblue]{hyperref}
\usepackage{xcolor}
\usepackage[table]{xcolor}

\usepackage{makecell}
\usepackage{multirow}
\usepackage{rotating}
\usepackage{lipsum}

\def\paperID{6} % *** Enter the Paper ID here
\def\confName{CVPR}
\def\confYear{2024}

\title{Advancing Cross-Domain Generalizability in Face Anti-Spoofing: \\Insights, Design, and Metrics}

\author{Hyojin Kim\textsuperscript{\rm 1}, Jiyoon Lee\textsuperscript{\rm 1,2}, Yonghyun Jeong\textsuperscript{\rm 1}, Haneol Jang\textsuperscript{\rm 3}, YoungJoon Yoo\textsuperscript{\rm 1,4}$^\dagger$ \\
\textsuperscript{\rm 1}Naver Cloud,\textsuperscript{\rm 2}Korea University, \textsuperscript{\rm 3}Hanbat National University, \textsuperscript{\rm 4}Chung-Ang University \\
{\tt\small hyojin.kimm@navercorp.com, jiyoonlee@korea.ac.kr, yonghyun.jeong@navercorp.com}, \\
{\tt\small hejang@hanbat.ac.kr, yjyoo3312@cau.ac.kr}
}

\begin{document}
\maketitle
\def\thefootnote{$\dagger$}\footnotetext{Corresponding author.}

\begin{abstract}
This paper presents a novel perspective for enhancing anti-spoofing performance in zero-shot data domain generalization.
Unlike traditional image classification tasks, face anti-spoofing datasets display unique generalization characteristics, necessitating novel zero-shot data domain generalization. 
One step forward to the previous frame-wise spoofing prediction, we introduce a nuanced metric calculation that aggregates frame-level probabilities for a video-wise prediction, to tackle the gap between the reported frame-wise accuracy and instability in real-world use-case.
This approach enables the quantification of bias and variance in model predictions, offering a more refined analysis of model generalization.
Our investigation reveals that simply scaling up the backbone of models does not inherently improve the mentioned instability, leading us to propose an ensembled backbone method from a Bayesian perspective. The probabilistically ensembled backbone both improves model robustness measured from the proposed metric and spoofing accuracy, and also leverages the advantages of measuring uncertainty, allowing for enhanced sampling during training that contributes to model generalization across new datasets.
We evaluate the proposed method from the benchmark OMIC dataset and also the public CelebA-Spoof and SiW-Mv2.
Our final model outperforms existing state-of-the-art methods across the datasets, showcasing advancements in Bias, Variance, HTER, and AUC metrics.
\end{abstract}    
\section{Introduction}
\label{sec:intro}

In response to the widespread adoption of deep learning in face recognition, detecting spoofing attacks like printed or video-displayed faces, also called presentation attacks, to the recognition system has become paramount for security.
Although sophisticatedly spoofed images often challenge human detection capabilities, the advances of recent face anti-spoofing (FAS) studies~\cite{zhang2020face,yu2023rethinking,sun2023rethinking,yu2020face,srivatsan2023flip} demonstrate that deep classification networks can effectively differentiate between authentic and counterfeit facial images. Nonetheless, developing a broadly generalized spoofing detection model, capable of accommodating diverse conditions, is still a challenging problem due to the inherent complications in spoofing datasets.

Specifically, each spoofing dataset possesses distinct attributes, stemming from varying data acquisition environments like camera capture conditions and backgrounds, introducing notable dataset biases. 
Given that visual cues for spoofing predominantly reside in the nuanced high-frequency image domain~\cite{chen2021generalized}, these biases significantly impede the extraction of reliable spoofing detection cues, thereby compromising the generalizability of FAS models. 
Recent Face Anti-Spoofing (FAS) research~\cite{shao2019multi,shao2020regularized,wang2022domain,liao2023domain,wang2022face,wang2023consistency,jia2020single,sun2023rethinking} employs evaluation metrics to showcase model generalizability over four benchmark anti-spoofing datasets, one step further to the cross dataset-domain adaptation (DA)~\cite{li2020face,zhou2022generative,huang2022adaptive,yue2023cyclically} for FAS problem. The domain generalization (DG) performance is evaluated by training on three of these datasets and subsequently evaluating the model's performance on the remaining one.
While the evaluation is a standard measure of FAS DG performance, real-world application of existing per-frame FAS methods to video inputs often yields unstable predictions across consecutive frames.

Our analysis begins with the observations as:
\begin{itemize}
\item Previous frame-wise FAS models successful for the existing FAS measurement fail to robustly capture the spoofness of the subsequent frames sharing similar semantic features.
\item Scaling up of the model size cannot effectively enhance the overall FAS performance including the robustness issue our first observations.
Such observations underscore the necessity of developing a novel metric to gauge model generalizability and a proper method to design a FAS model, and second, effective model design to improve FAS performance beyond scaling up the model size.
\end{itemize}

In this paper, we present a novel perspective on a paradigm for evaluating and designing FAS models. 
First, we introduce a novel evaluation metric, variance, and bias, focusing on temporal coherence and noise resilience for robustness assessment. Our frame-level analysis validates these metrics, providing quantitative evidence supporting the instability of prior FAS methods.
Second, we highlight the efficacy of ensemble methodologies in enhancing spoofing detection performance for both the previous and the newly introduced evaluation protocols. Through extensive empirical evaluations, we demonstrate that the ensemble strategy applying Monte-Carlo (MC) dropout~\cite{gal2016dropout} is a pivotal design option for making a FAS model that enhances precise spoofing detection and superior generalizability.
Last, through empirical analyses to determine the optimal backbone for MC-dropout-based ensembling, we introduce ECLIPS, a FAS model based on the CLIP vis encoder~\cite{radford2021learning}. Our proposed ECLIPS achieves state-of-the-art FAS detection performance while maintaining frame-level prediction consistency.
It is worth noting that, instead of processing consecutive video frames, which incurs high computational costs, we demonstrate the efficacy of the MC-dropout-based ensembling method in improving frame-level FAS prediction consistency.

In demonstrating the effectiveness of our proposed approach, we conduct comprehensive qualitative and quantitative experiments across four benchmark FAS datasets: OULU~\cite{OULU} (\textbf{O}), CASIA~\cite{zhang2012face} (\textbf{C}), Idiap~\cite{Chingovska} (\textbf{I}), MSU-MFSD~\cite{Diwen2014} (\textbf{M}), and additionally to the CelebA-Spoof~\cite{CelebA-Spoof} and SiW-Mv2~\cite{liu2019deep} datasets. 
The experimental results demonstrate that our ensemble-based FAS model ECLIPS outperforms current state-of-the-art FAS models by a notable margin including recent FAS methods leveraging auxiliary multi-modality both for FAS accuracy and for the frame-level prediction robustness validated through the proposed protocol.

In summary, we summarize the contributions of the proposed method as follows:
\begin{itemize}
\item We provide observations for real-world FAS applications for video input, with a focus on frame-level prediction robustness and model scalability, while introducing the remaining challenges in current FAS research.
\item We introduce temporal and noise-aware robustness metrics, by calculating variance and bias, demonstrating their potential as reliable indicators for ensuring the generalizability of FAS models.
\item Through empirical analyses, we highlight the proposed ensemble approach as a potent design strategy for FAS models. Consequently, our proposed FAS model ECLIPS reports state-of-the-art performance, excelling in FAS generalization capabilities. 
\end{itemize}

\section{Related Work}
\label{sec: related work}

\paragraph{Domain Generalization for FAS}
To deal with the well-known dataset disparities inherent to the FAS, many FAS studies have been proposed, emphasizing cross-dataset domain (cross-domain) evaluations via domain adaptation (DA)~\cite{li2020face,zhou2022generative,huang2022adaptive,yue2023cyclically}, domain generalization (DG)~\cite{shao2019multi,shao2020regularized,wang2022domain,liao2023domain,wang2022face,wang2023consistency,jia2020single,sun2023rethinking}, or both~\cite{srivatsan2023flip}.
While the DA paradigm allows for few-shot adaptation to the target domain, DG offers a more rigorous assessment by gauging zero-shot generalization across datasets. Given real-world scenarios where DA might be infeasible, the DG approach gains prominence due to its straightforward applicability.
After the initiative DG approach~\cite{shao2019multi} proposing Mult-adversarial domain generalization, many follow-up studies have been proposed by meta-laerning~\cite{shao2020regularized}, style-shuffling~\cite{wang2022domain}, new losses~\cite{liao2023domain,wang2023consistency,sun2023rethinking} based on dataset analyses, and multi-modaility~\cite{srivatsan2023flip}.
These studies typically evaluate the generalization ability of their methods by using four representative benchmark datasets: OULU~\cite{OULU}, CASIA~\cite{zhang2012face}, Idiap~\cite{Chingovska}, and MSU-MFSD~\cite{Diwen2014}, leaving one dataset for evaluation from the model trained by the other three datasets.
Advancing beyond the previous work, we demonstrate that even with consistent improvements in conventional evaluation frameworks, the reported outcomes might not assuredly reflect detection proficiency across diverse datasets, as evidenced by our newly introduced FAS dataset.
  
\paragraph{Designing Principles for FAS}
Followed by the improvement of deep classification architecture~\cite{resnet,dosovitskiy2020image}, we have observed consistent enhancement in the detection capability of the presentation attack (PA) samples by using the classification architecture as their backbone.
Most previous FAS methods~\cite{shao2020regularized,yang2020pipenet,wang2022domain,liao2023domain,wang2023consistency,sun2023rethinking,patchnet} have configured the PA detection model as a form of classification network using ImageNet~\cite{deng2009imagenet} pre-trained ResNet~\cite{resnet}-like architecture.
Instead of scaling up the backbone, FAS methods have focused on the data domain-specific improvements, such as analyzing the effect of input configuration~\cite{patchnet,yang2020pipenet} of feature space manipulation~\cite{wang2023consistency,sun2023rethinking}, based on the fixed ResNet backbone.
A notable size and performance scale-up of the architecture is followed by the use of vision transformer (ViT)~\cite{dosovitskiy2020image} pre-trained by large-scale image and text multi-modal dataset~\cite{schuhmann2022laion} as CLIP visual encoder. Followed by the initiative attempts~\cite{george2021effectiveness,liao2023domain} using ImageNet-pre-trained ViT for FAS, adoption of ViT pre-trained as CLIP~\cite{radford2021learning} visual encoder reported impressive performance enhancement~\cite{srivatsan2023flip}.

\begin{table*}[t]
\centering
\begin{tabular}{c|cccccc|ccc}
\hline
Sample No. & \multicolumn{6}{c|}{Frame-wise} & \multicolumn{3}{c}{Video-wise} \\ \hline
GT Label = 1 & Frame 1 &  Frame 2 &  Frame 3 &  Frame 4 &  Frame 5 & Avg. Prob. & Prob. & Bias & Variance \\ \hline
 ResNet18~\cite{resnet}& 0.97(1) & 0.33(0) & 0.52(1) & 0.17(0) & 0.98(1) & 0.59(1) & 0.79(1) & 0.044 & 0.40 \\
 ViT-B/16~\cite{dosovitskiy2020image} & 0(0) & 0.55(1) & 0.98(1) & 0(0) & 0.17(0) & 0.4(0) & 0.34(0) & 0.435 & 0.31 \\
 FLIP~\cite{srivatsan2023flip} & 0.93(1) & 0.98(1) & 0.96(1) & 0.54(1) & 0.82(1) & 1.0(1) & 0.84(1) & 0.025 & 0.14 \\
 ECLIPS & 0.99(1) & 0.98(1) & 0.97(1) & 0.88(1) & 0.99(1) & 1.0(1) & 0.96(1) & 0.001 & 0.05 \\
\hline
\end{tabular}
\caption{Quantitative analysis of prediction variability in frame-by-frame images from commonly used backbone models~\cite{resnet,dosovitskiy2020image}. The numerical values within frame-wise indicate the prediction probabilities for each frame, where the numbers in parentheses represent a decision of spoofing (0) or live (1). In the video-wise results, the average probability (Prob.) of all frames is presented, with the overall video prediction denoted within the parentheses. Additionally, the results include calculations of bias and variance.}

\label{table:robustness}
\end{table*}
In this paper, we investigate the effect of backbone scale-up in various aspects including depth, channel, and pre-trained dataset on DG scenario accompanied by the novel robustness evaluation metrics and show the efficacy of the ensembling-based method. Beyond previous attempts~\cite{muhammad2023deep,jiang2022depth} using the ensembling method for frame-skipping~\cite{muhammad2023deep} and integrating data-domain specific models~\cite{jiang2022depth},  the ensemble approach as a pivotal mechanism for theoretically addressing DG in the FAS, underpinned by foundational FAS insights and corroborative analyses.

\section{Observations}
In this section, we initiate our discussion of designing a generalizable FAS model grounded in observations from FAS evaluations concerning \textit{temporal robustness} and the \textit{scalability} of the prediction. To quantitatively measure the temporal robustness, we introduce the terms Variance and Bias. 
\subsection{Temporal Robustness}
\label{sec:temporal_robustness}
\begin{figure}[t]
    \centering
    \includegraphics[width=1.0\linewidth]{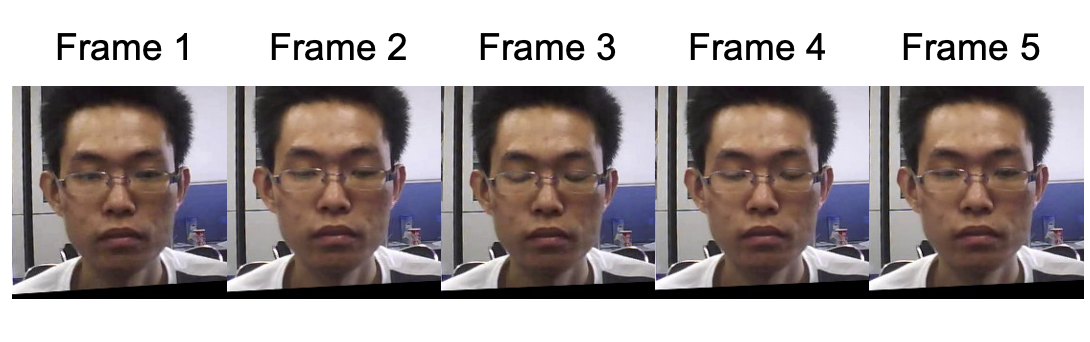}
    \caption{Visualization of images containing a sequence of five frames extracted from the video within the CASIA~\cite{zhang2012face}. The images demonstrate a range of facial expressions and pose variations.}
    \label{fig:enter-label}
\end{figure}

As shown in Figure~\ref{fig:enter-label}, the provided image shows a sequence of five frames from a video, labeled as Frame 1 through Frame 5. Each frame features the same individual, capturing subtle changes in facial expression and pose, which are critical for evaluating the consistency and robustness of face anti-spoofing models across different moments in time. This sequence demonstrates how frame-wise analysis can reveal variations in the model's prediction accuracy, thus emphasizing the importance of a model's ability to maintain stable performance throughout a video sequence. Table~\ref{table:robustness} provides a quantitative analysis corresponding to the images shown in Figure~\ref{fig:enter-label}. The figure illustrates sequence moments within the same video, while Table~\ref{table:robustness} evaluates the performance of various models at each respective moment.

The computational expressions for these methodologies are outlined as follows:
Frame-wise probability refers to the probability score assigned to an individual frame, indicating whether the frame is likely live or a spoof. Frame-wise prediction is expressed verbally as the average of outcomes where each outcome is 1 if the probability score of the frame exceeds a certain threshold and 0 otherwise. This translates to a binary classification where scores above the threshold indicate a live frame and scores below indicate a spoof. Video-wise probability is explained as the average of the probability scores of all frames within a video being compared against a threshold to decide the overall nature of the video. Video-wise Prediction involves making a binary decision, which is determined as live (1) if the average probability score of all frames exceeds the threshold and as spoof (0) if it does not. 

\begin{table*}
\centering
\resizebox{\textwidth}{!}{% 
\begin{tabular}{lllccccccccccc}
\Xhline{1pt}
\Gape[12pt][6pt]{\multirow{2}{*}{Method (\%)}} &
  \multirow{2}{*}{FAS} &
  \multirow{2}{*}{\begin{tabular}[c]{@{}l@{}}Backbone\\ Architecture\end{tabular}} &
  \multirow{2}{*}{MBParams} &
  \multirow{2}{*}{GFLOPs} &
  \multirow{2}{*}{\begin{tabular}[c]{@{}l@{}}Size or \\ Latency\end{tabular}} &
  \multicolumn{2}{c}{OCI$\rightarrow$M} &
  \multicolumn{2}{c}{OMI$\rightarrow$C} &
  \multicolumn{2}{c}{OCM$\rightarrow$I} &
  \multicolumn{2}{c}{ICM$\rightarrow$O} \\
                                  &                   &              &  &  &  & HTER ($\downarrow$) & AUC ($\uparrow$) & HTER ($\downarrow$) & AUC ($\uparrow$) & HTER ($\downarrow$) & AUC ($\uparrow$) & HTER ($\downarrow$) & AUC ($\uparrow$) \\ \Xhline{1pt}
\multirow{3}{*}{ResNet \cite{resnet}}         & \multirow{3}{*}{}           & ResNet18     & 42.63 & 2.38 & 2.51 & 10.24 & 96.15  & 14.72 & 92.76 & 17.00 & 91.72 & 15.03 & 92.34 \\
                                  &                                                   & ResNet34     & 81.19 & 4.80 & 3.89 &  \textbf{8.57} & \textbf{96.22} & \textbf{11.93} & \textbf{94.51} & 18.00 & 91.13 & \textbf{13.11} & 93.31 \\
                                  &                                                   & ResNet50     & 89.68 & 5.40 & 7.06 & 10.24 & 93.97 & 13.94 & 93.97 & \textbf{12.95} & \textbf{94.46} & 13.70 & \textbf{93.70} \\ \hline
\multirow{3}{*}{EfficientNet-V2 \cite{tan2021efficientnetv2}}   & \multirow{3}{*}{}           & Tiny      & 47.71 & 2.50 & 19.19 & 11.67	& 94.24 & 16.16	& 92.64 & 20.45	& 83.42 & 16.09	& 91.84    \\
                                  &                                                   & Small     & 83.88 & 3.84 & 20.47 & \textbf{8.81} & \textbf{95.86} & \textbf{12.71} & \textbf{93.12} & 22.50 & 86.46 & \textbf{15.91} & \textbf{92.00}    \\
                                  &                                                   & Medium    & 193.70 & 8.08 & 30.87 & 12.86	& 94.99 & 22.50	& 78.87 & \textbf{14.83} & \textbf{93.48} & 17.95 & 90.16\\ \hline
\multirow{3}{*}{ViT \cite{dosovitskiy2020image}}            & \multirow{3}{*}{}           & Tiny/16  & 20.93 & 1.41 & 5.40 & 11.43	& 95.68 &  15.94	& 91.16  & 21.90	& \textbf{88.94} & \textbf{16.07}	& 90.82     \\
                                  &                                                   & Small/16 & 82.36 & 5.54 & 5.45 &  \textbf{8.57} & 96.08  &  14.72	& 92.77  &  \textbf{20.00} & 85.32  & 14.44 & \textbf{91.61}     \\
                                  &                                                   & Base/16  & 326.72 & 22.00 & 6.31 &  10.24	& \textbf{96.10}  &  \textbf{13.94}	& \textbf{93.91}  &  21.00 & 78.32 & 17.99	& 89.38    \\ \hline
\multirow{3}{*}{SAFAS \cite{sun2023rethinking}}          & \multirow{3}{*}{\checkmark} & ResNet18     & 42.63 & 2.38 & 2.51 &  10.24	& \textbf{95.97}  & \textbf{12.04} & \textbf{94.28} &  \textbf{8.90} & 96.96 & 11.40 & 95.08     \\
                                  &                                                   & ResNet34    & 81.19 & 4.80 & 3.89 &  \textbf{8.33}	& 95.94  &  12.04 & 94.28  & 9.05	& \textbf{97.79}  &  \textbf{10.24} & \textbf{95.99}     \\ 
                                  &                                                   & ResNet50     & 89.68 & 5.40 & 7.06 & 11.19 & 95.53 & 13.38	& 94.19&  10.50 & 96.31 & 11.25 & 95.56 \\ \hline
\multirow{3}{*}{PatchNet \cite{patchnet}}       & \multirow{3}{*}{\checkmark} & ResNet18     & 42.63 & 2.38 & 7.06 & \textbf{10.24}	& 93.9 & 16.05	& 90.05 & 22.00 & 88.12 & \textbf{17.22} & 88.53 \\ 
                                  &                                                   & ResNet34     & 81.19 & 4.80 & 3.89 &  11.67	& 95.13  &15.94	&89.33  & \textbf{14.05}	& \textbf{93.48} & 17.80 & \textbf{88.90} \\
                                  &                                                   & ResNet50     & 89.68 & 5.40 & 7.06 & 11.43	& \textbf{95.85} & \textbf{13.38} & \textbf{93.25} & 19.05	&90.03 &18.10	&88.26\\ \hline

\end{tabular}}
\caption{Comparative experiment of FAS methodologies for \textbf{HTER} (Half Total Error Rate) and \textbf{AUC} (Area Under the Curve) with respect to model scale variation. The overall quantitative results show that scaling-up of the model size does not yield substantial improvements in FAS applications.}
\label{tab:scale-h}
\end{table*}

Concerning bias and variance within these metrics, they are defined as follows: 
\begin{itemize}
\item {\textbf{Bias} $B(\cdot)$ is quantified by the mean squared error (MSE) between the actual labels and the video-wise probability scores. This measures how accurately the video-wise predictions reflect the true outcomes. 
\begin{eqnarray}
\begin{aligned}
B(Y_i) = \frac{1}{N} \sum_{i=1}^{N} (Y_i - \hat{Y}_i)^2.
\label{eq:bias}
\end{aligned}
\end{eqnarray}
In this formula, $\mathcal{N}$ represents the total number of videos, $Y_i$ is the actual label for the $i$th video (live or spoof), and $\hat{Y}_i$ is the predicted video-wise probability for the $i$th video.

} 

\item {\textbf{Variance} is calculated as the average of the standard deviations of the frame-wise probability scores across all videos. This standard deviation reflects the consistency of frame-wise probability scores within a video, providing insights into the model's reliability across different frames.
\begin{eqnarray}
\begin{aligned}
V(P_i) = &\frac{1}{N} \sum_{i=1}^{N} \sigma^2(P_i),\\
\sigma^2(P_i) = &\frac{1}{M_i} \sum_{j=1}^{M_i} (P_{ij} - \bar{P}_i)^2.
\label{eq:varandsigma}
\end{aligned}
\end{eqnarray}
The Variance $V(\cdot)$ is computed as the mean of the standard deviations (denoted by $\sigma$) of frame-wise probability scores across all videos. $M_{i}$ denotes the total number of frames in the $i$th video, $P_{ij}$ represents the probability score of the $j$th frame in the $i$ video, and $\bar{P}_{i})$ is the mean of the frame-wise probability scores for the $i$th video.
}
\end{itemize}

This distinction between frame-wise and video-wise methodologies not only highlights their unique analytical perspectives but also underlines the complexities involved in assessing the reliability and precision of video-based live detection systems. Through the evaluation of bias and variance, a deeper understanding of the model's performance is achieved, allowing for a balance between accuracy and consistency across various videos and frames.

Table~\ref{table:robustness} quantitatively illustrates the model's performance in terms of bias and variance across individual frames within a single video, highlighting the disparity in frame-wise predictions. For each model, The Frame-wise section in Table~\ref{table:robustness} displays the probability scores (with predicted labels in parentheses) for five representative frames and the average probability of prediction (Avg. Prob.) across these frames. The Video-wise section further quantifies the overall prediction probability (Prob.), bias, and variance, offering insights into the model's stability and generalization capability. Notably, the our proposed model demonstrates remarkably low variance in frame-wise predictions, signifying consistent performance across the video's duration, which is crucial for reliable face anti-spoofing in dynamic scenarios.
 \begin{figure*}
    \centering
    \includegraphics[width=0.90\linewidth]{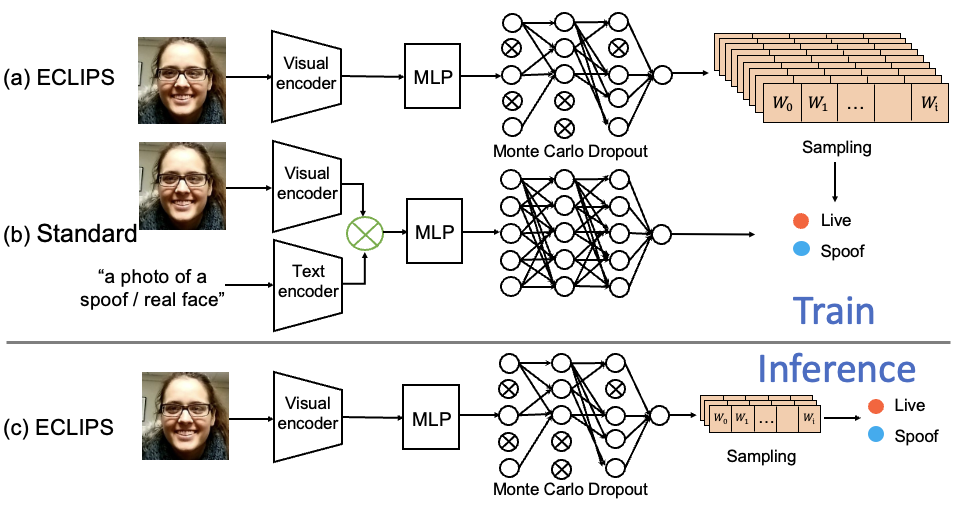}
    \vspace{-2mm}
    \caption{Architecture of the proposed FAS Model ECLIPS. (a) The ECLIPS model for training utilizes only a Visual Encoder. (b) The Standard variant is a dual-stream model that integrates textual information with visual features. (c) The ECLIPS, at inference, only utilizes a Visual Encoder.}
    \label{fig:overview}
\end{figure*}

\subsection{Scalability}
\label{sec:scalability}
Although it is generally observed that larger models tend to exhibit higher performance, there are scenarios where this does not hold true.
Given the datasets commonly used in the FAS task (OULU~\cite{OULU}, CASIA~\cite{zhang2012face}, Idiap~\cite{Chingovska}, MSU-MFSD~\cite{Diwen2014}), characterized by its relatively small scale, it becomes imperative to validate performance across various backbone scales. This necessity stems from the nuanced understanding that while larger models have the potential to achieve higher accuracies, their effectiveness can be constrained by factors such as overfitting, increased computational requirements, and the curse of dimensionality, especially in the context of limited data availability. Therefore, we conducted experiments to validate the performance in terms of the scalability of various models. As seen in 
Table~\ref{tab:scale-h} illustrates the performance variations according to the scale of models well-utilized in the FAS problem, including ResNet~\cite{resnet}, EfficientNet-V2~\cite{tan2021efficientnetv2}, and ViT~\cite{dosovitskiy2020image}, in conjunction with anti-spoofing applications of contrastive learning~\cite{sun2023rethinking} and patch-based learning~\cite{patchnet} methods. Through these results, we confirm that irrespective of the type of model and learning method, the impact of the model's scale on training performance is negligible.

\section{ECLIPS}
\label{sec: proposed method}
Guided by the overall insights, we propose a temporal and noise-aware robustness framework called ECLIPS for cross-domain FAS. The proposed framework is comprised of two main components: a base learner module and a decision fusion module, as illustrated in Figure~\ref{fig:overview}. We describe each component in the following sections.

\subsection{Base Learner Module}
Compared to datasets used in other vision tasks, commonly used datasets for face anti-spoofing, such as OULU~\cite{OULU}, CASIA~\cite{zhang2012face}, Idiap~\cite{Chingovska}, and MSU-MFSD~\cite{Diwen2014}, have a notably smaller scale. This limited data scope can cause models to overfit, which in turn reduces their ability to generalize across different domains. Simply increasing the scale of the model is not an effective alternative when data is scarce. 

To address this issue, we employ ensemble techniques that leverage multiple base learners to capture diverse domain characteristics. Our goal is to assess the effectiveness of the ensemble approach by comparing models with the same backbone architecture as base learners, irrespective of the backbone model used. Ultimately, we determine that the CLIP~\cite{radford2021learning} is highly effective for use as a base learner. In the proposed fusion learning, we adopt the CLIP as our backbone base learner for tackling the FAS challenge. 

Despite the constraints posed by the limited FAS training data, capturing its inherent diversity during the CLIP~\cite{radford2021learning} training process is crucial for robustness against emerging attack types. We adopt diverse pre-training strategies for each base learner at the data and model levels to address these novel attacks effectively. In the final base learner, we use the MC dropout \cite{gal2016dropout} technique at the model level to improve the generalization ability. 

\subsection{Decision Fusion Module}
Ensemble learning trains multiple base learners and aggregates their outputs using specific rules. While many ensemble models focus on the architecture and use averaging to predict outputs, this simplistic averaging often leads to suboptimal performance, lacking adaptability to data and sensitivity to biases in base learners. Errors, especially from overfitting in deep learning architectures, can further exacerbate this issue.

To address these challenges, we adopt an approach where we learn weights for each model, constructing a decision fusion module to aggregate probability values from each model for the final prediction. By learning these weights, the fusion module reflects the contributions of each base learner, promoting diversity while assigning higher weights to more important models.

Figure~\ref{fig:overview} illustrates the architecture of the ECLIPS model, seamlessly integrating a CLIP Visual Encoder and a Multilayer Perceptron (MLP) with Monte Carlo Dropout for uncertainty estimation. During training, the CLIP Visual Encoder processes image inputs, followed by applying dropout with a probability of 0.5 to the extracted features. These dropout-applied features are then used for generating predictions through sampling, iterating this process 10 times to capture uncertainty and combat overfitting. During inference, the process is streamlined for efficiency without compromising robustness, with only 3 samplings performed, ensuring accurate predictions in real-time scenarios.

\subsection{Implementation Details}
In the preprocessing phase, the Multi-Task Cascaded Convolutional Networks (MTCNN)~\cite{zhang2016joint} was used to identify facial bounding boxes. These boxes were then expanded by a padding of 0.5 to crop the faces. After this step, we used a total of $M=32$ frames each frame from the real and fake videos into the model. In our architecture, we incorporated a feature embedding layer with a dropout layer, setting the dropout rate at 0.5, and utilized a CLIP Visual encoder based on the ViT-B/16 as the backbone. 

\begin{table*}
\centering
\resizebox{\textwidth}{!}{% 

\begin{tabular}{lcccccccc|cc}
\Xhline{1pt}
\Gape[12pt][6pt]{\multirow{2}{*}{Method (\%)}} &
  \multicolumn{2}{c}{OCI$\rightarrow$M} &
  \multicolumn{2}{c}{OMI$\rightarrow$C} &
  \multicolumn{2}{c}{OCM$\rightarrow$I} &
  \multicolumn{2}{c|}{ICM$\rightarrow$O} &
  \multicolumn{2}{c}{Average} \\
                                  &  HTER ($\downarrow$) & AUC ($\uparrow$) & HTER ($\downarrow$) & AUC ($\uparrow$) & HTER ($\downarrow$) & AUC ($\uparrow$) & HTER ($\downarrow$) & AUC ($\uparrow$) & HTER ($\downarrow$) & AUC ($\uparrow$) \\ \Xhline{1pt}
MADDG (CVPR' 19) \cite{shao2019multi}            & 17.69 & 88.06 & 24.50 & 87.51 & 22.19 & 84.99 & 27.98 & 80.02 & 23.09 & 85.89 \\
MDDR (CVPR' 20) \cite{wang2020cross}             & 17.02 & 90.10 & 19.68 & 87.43 & 20.87 & 86.72 & 25.02 & 81.47 & 20.64 & 86.43 \\
NAS-FAS (TPAMI' 20) \cite{yu2020fas}             & 16.85 & 90.42 & 15.21 & 92.64 & 11.63 & 96.98 & 13.16 & 94.18 & 14.21 & 93.80 \\
RFMeta (AAAI'20)SAFAS \cite{shao2020regularized} & 13.89 & 93.98 & 20.27 & 88.16 & 17.30 & 90.48 & 16.45 & 91.16 & 16.97 & 90.69 \\
$D^2$AM (AAAI'21) \cite{chen2021generalizable}   & 12.70 & 95.66 & 20.98 & 85.58 & 15.43 & 91.22 & 15.27 & 90.87 & 16.09 & 90.58 \\
DRDG(IJCAI '21) \cite{liu2021dual}               & 12.43 & 95.81 & 19.05 & 88.79 & 15356 & 91.79 & 15.63 & 91.91 & 15.66 &  91.82 \\
self-DA (AAAI' 21) \cite{wang2021self}           & 15.40 & 91.80 & 24.50 & 84.40 & 15.60 & 90.10 & 23.10 & 84.30 & 19.65 & 87.15 \\
ANRL (ACM MM' 21) \cite{liu2021adaptive}         & 10.83 & 96.75 & 17.85 & 89.26 & 16.03 & 91.04 & 15.67 & 91.90 & 15.09 & 92.23 \\
FGHV (AAAI' 21) \cite{liu2022feature}            & 9.17  & 96.92 & 12.47 & 93.47 & 16.29 & 90.11 & 13.58 & 93.55 & 12.87 & 93.51 \\
SSDG-R (CVPR' 20) \cite{jia2020single}           & 7.38  & 97.17 & 10.44 & 95.94 & 11.71 & 96.59 & 15.61 & 91.54 & 11.28 & 95.06 \\
SSAN-R (CVPR' 22) \cite{wang2022domain}          & 6.67  & 98.75 & 10.00 & 96.67 & 8.88  & 96.79 & 13.72 & 93.63 & 9.80 & 96.21 \\
PatchNet (CVPR' 22) \cite{patchnet}       & 7.10  & 98.46 & 11.33 & 94.58 & 13.40 & 95.67 & 11.82 & 95.07 & 10.90 & 95.19 \\
GDA (ECCV' 22) \cite{zhou2022generative}         & 9.20  & 98.00 & 12.20 & 93.00 & 10.00 & 96.00 & 14.40 & 92.60 & 11.45 & 94.65 \\
SA-FAS (CVPR' 23) \cite{sun2023rethinking}       & 5.95  & 96.55 & 8.78  & 95.37 & 6.58  & 97.54 & 10.00 & 96.23 & 7.83 & 96.17 \\
DIVT-M (WACV' 23) \cite{liao2023domain}          & 2.86  & 99.14 & 8.67  & 96.62 & 3.71  & 99.29 & 13.06 & 94.04 & 7.07 & 97.27 \\
ViT (ECCV' 22) \cite{huang2022adaptive}          & 1.58  & 99.68 & 5.70  & 98.91 & 9.25  & 97.15 & 7.47  & 98.42 & 6.00 & 98.04 \\ \Xhline{1pt}
FLIP-MCL (ICCV' 23) w/ViT-B/16 \cite{srivatsan2023flip}       & 4.95  & 98.11 & 0.54  & 99.98 & 4.25  & 99.07 & 2.31  & 99.63 & 3.01 & 99.19 \\
FLIP-MCL (ICCV' 23) w/ViT-L/14 \cite{srivatsan2023flip}       & 4.52  & 98.40 & 0.11  & 99.99 & 3.45  & 99.50 & 1.81  & 99.63 & 2.47 & 99.13 \\ \Xhline{1pt} \rowcolor{green!10}
ECLIPS (w/Monte Carlo Dropout)                                   & 1.43  & 99.87 & 0.78  & 99.18 & 1.60  & 99.16 & 3.21  & 99.35 & \textbf{\textcolor{blue}{1.76}} & \textbf{\textcolor{blue}{99.39}}  \\ \Xhline{1pt}

\end{tabular}}
\caption{Quantitative FAS evaluation result below HTER and AUC metrics. Across transfer learning protocols OMIC, it achieved the lowest average HTER and the highest AUC scores. This suggests a superior efficacy of the OMIC protocol in enhancing the robustness and reliability of FAS systems against spoofing attempts, underscoring its potential for integration into advanced biometric authentication frameworks. Denoted in blue, confirming its status as the new state-of-the-art in the field.}
\label{tab:tab4}
\end{table*}
To establish the final logit for a classifier layer, we sample the logit values and average these samples. For the training setup, an input size of 224x224 and a batch size of 16 over 100 epochs for the learning. For optimization purposes, the ADAM optimizer is employed, configured with a learning rate of $1e^{-6}$ and a weight decay to $1e^{-6}$. 
\section{Experiments}
\label{sec: experiments}

\subsection{Experimental Setup}
\paragraph{Dataset and Domain Generalization Protocols}

We evaluated the proposed FAS model using both the public and newly presented FAS datasets.
First, followed by the previous domain generalization (DG) setting, we use for the public datasets OULU~\cite{OULU} (\textbf{O}), CASIA~\cite{zhang2012face} (\textbf{C}), Idiap~\cite{Chingovska} (\textbf{I}), and MSU-MFSD~\cite{Diwen2014} (\textbf{M}) for the evaluation.
In the DG scenario, the training and test set consists of different sets of datasets such as the training dataset pairs of \textbf{O}, \textbf{C}, and \textbf{I} datasets and the test dataset as M, abbreviated as (\textbf{OCI$\rightarrow$M}).  In addition to the previous setting, we use public FAS datasets: CelebA-Spoof~\cite{zhang2021celebaspoof} (\textbf{A}) and SiW-Mv2~\cite{guo2022multi} (\textbf{S}), to show the tendency the existing DG evaluation, as (\textbf{OCIM$\rightarrow$\textbf{A})}. From the newly introduced setting, we measure the credibility of the previous DG evaluation by analyzing the correspondence between the two test datasets, CelebA-Spoof and SiW-Mv2. From this setting, we aim to show both the superiority of the proposed FAS method and also the DG representation credibility of the proposed robustness-aware evaluation metric.

\paragraph{Evaluation Metrics} 
First, following the conventional settings of \cite{sun2023rethinking,srivatsan2023flip}, we evaluate the DG performance using metrics that have been previously established: Half Total Error Rate (HTER), Area Under the Receiver Operating Characteristic Curve (AUC), and  True Positive Rate (TPR) at a fixed False Positive Rate (FPR). 
Additionally, we assess the robustness of the FAS method by introducing measures of Variance and Bias in Section~\ref{sec:temporal_robustness}. These metrics are designed to quantify robustness in terms of temporal stability and sensitivity to noise.

\subsection{Cross-Domain Performance}
In this section, we compare the performance of models using the OCIM benchmark, designed to assess traditional cross-domain performance. The OCIM benchmark experiments, conducted with a leave-one-out protocol, include four transfer scenarios denoted as OCI→M, OMI→C, OCM→I, and ICM→O.  In our comprehensive study, which aims to enhance FAS methodologies, we evaluate our model's performance not only through traditional metrics like HTER and AUC but also by proposing the inclusion of novel generalization metrics such as bias and variance.
\paragraph{Traditional Evaluation}

\begin{table*}
\centering
\resizebox{\textwidth}{!}{% 
\begin{tabular}{lcccccccc|cc}
\Xhline{1pt}
% \vspace{0.3cm}
% \Gape[12pt][6pt]{\multirow{2}{*}{Method (\%)}} &
\multirow{2}{*}{Method (\%)} &
  \multicolumn{2}{c}{OCI$\rightarrow$M} &
  \multicolumn{2}{c}{OMI$\rightarrow$C} &
  \multicolumn{2}{c}{OCM$\rightarrow$I} &
  \multicolumn{2}{c|}{ICM$\rightarrow$O} &
  \multicolumn{2}{c}{Average} \\
                                  &  Bias ($\downarrow$) &  Variance ($\downarrow$) &  Bias ($\downarrow$) & Variance ($\downarrow$) &  Bias ($\downarrow$) & Variance ($\downarrow$) &  Bias ($\downarrow$) & Variance ($\downarrow$) & Bias ($\downarrow$) & Variance ($\downarrow$)  \\ 
\Xhline{1pt}
ResNet18 \cite{resnet}                & 0.085 & 0.081 & 0.080 & 0.128 & 0.094 & 0.055 & 0.258 & 0.127 & 0.129 & 0.097 \\
XceptionNet \cite{chollet2017xception}             & 0.088 & 0.085 & 0.076 & 0.141 & 0.109 & 0.114 & 0.295 & 0.149 & 0.142 & 0.122\\
EfficientNet-V2/Tiny \cite{tan2021efficientnetv2}    & 0.078 & 0.093 & 0.075 & 0.129 & 0.085 & 0.104 & 0.319 & 0.146 & 0.139 & 0.118  \\
ViT-B/16 \cite{dosovitskiy2020image}                & 0.084 & 0.072 & 0.075 & 0.126 & 0.075 & 0.080 & 0.131 & 0.135 & 0.091 & 0.103 \\
SA-FAS (CVPR' 23) \cite{sun2023rethinking}       & 0.071 & 0.085 & 0.159 & 0.131 & 0.064 & 0.107 & 0.231 & 0.122 & 0.131 & 0.111 \\
FLIP-MCL (ICCV' 23) \cite{srivatsan2023flip}        & 0.025 & 0.064  & 0.029 & 0.077 & 0.054 & 0.121 & 0.152 & 0.143 & 0.065 & 0.101 \\ \Xhline{1pt} \rowcolor{green!10} 
ECLIPS & 0.021 & 0.075 & 0.028 & 0.098 & 0.039 & 0.076 & 0.051 & 0.054 & 0.034 & 0.075 \\ 
\Xhline{1pt}
\end{tabular}}
\caption{Comparative Evaluation of ECLIPS against state-of-the-art models across four transfer scenarios on bias and variance metrics. This table showcases the superior performance of our ECLIPS model, demonstrating the lowest average bias and variance among all tested models, thereby establishing a new benchmark in the field of FAS.}
\label{tab:scal-v}
\end{table*}
\begin{table*}
\centering
\resizebox{0.90\linewidth}{!}{% 
\begin{tabular}{ll|cc|cccc}
% \Xhline{1pt}
\hline
\Gape[12pt][6pt]{\multirow{2}{*}{Method (\%)}} &
% \multirow{2}{*}{Method (\%)} &
  \multirow{2}{*}{GFLOPs} &
  \multicolumn{2}{c|}{OCIM$\rightarrow$A} & 
  \multicolumn{4}{c}{OCIM$\rightarrow$S} \\
                                                            &  & HTER ($\downarrow$) & AUC ($\uparrow$) & HTER ($\downarrow$) & AUC ($\uparrow$) & Bias ($\downarrow$) & Variance ($\uparrow$) \\ 
                                                            \hline
                                                            % \Xhline{1pt}
ViT-B/16 \cite{dosovitskiy2020image}                             & 22.00 & 38.64 & 66.09 & 44.82 & 58.12 & 0.44 & 0.04 \\  
ViT-B/16 (w/Sampling)              & 66.00 & 20.46 & 83.82 & 29.52 & 70.23 & 0.31 & 0.16 \\  
ViT-B/16 (w/Monte Carlo Dropout)  & 66.00 & 17.76 & 89.52 & 28.92 & 79.80 & 0.20 & 0.16  \\  \Xhline{1pt}
CLIP-V \cite{radford2021learning}                          & 11.27 & 13.05 & 92.46 & 30.27 & 77.51 & 0.24 & 0.15 \\  
CLIP-V (w/Sampling)            & 33.81 & 8.53 & 95.57 & 27.57 & 80.38 & 0.23 & 0.15  \\  
FLIP-MCL (ICCV' 23) \cite{srivatsan2023flip}                  & 103.57 & 10.73 & 94.86 & 28.54 & 79.49 & 0.26 & 0.08 \\ \Xhline{1pt}  \rowcolor{green!15}
ECLIPS                                                  & 33.81 & 8.47  & 96.16 & 26.70 & 81.29 & 0.18 & 0.14 \\ \Xhline{1pt}

\end{tabular}}
\caption{Evaluation of domain generalization for models trained on OCIM to new datasets. The CelebA-Spoof and SiW-Mv2 datasets encompass a vast array of spoof types, surpassing the scope of the original OCIM dataset. Notably, CelebA-Spoof comprises data without accompanying videos, thus excluding Variance and Bias considerations.
}
\label{tab:tab6}
\end{table*}
Table~\ref{tab:tab4} presents a refined evaluation of FAS models on the OCIM dataset, demonstrating superior HTER and AUC metrics. Leveraging the FLIP-MCL~\cite{srivatsan2023flip} as a baseline, we have developed a model that not only achieves greater computational efficiency but also improves performance and generalization. Furthermore, by incorporating innovative fusion techniques such as Monte Carlo Dropout, our model gains additional advantages in robustness and reliability, as evidenced in the reported results. Importantly, our research highlights the considerable benefits of using an ensemble of models over a single model architecture. This ensemble approach enhances the representational capacity of our system and increases its resilience against various spoofing attacks, thereby improving its generalization across unseen data domains. Ultimately, our model achieves state-of-the-art performance with fewer parameters compared to the existing FLIP-MCL Large model.
Our experimental findings underline the advantages of employing a constellation of smaller backbones over a single larger one to boost model generalization, aligning with the paradigm shift suggested by our initial experiments.

\paragraph{Temporal Robustness Evaluation}

This chapter provides comparison results for variance and bias, designed to consider the model's confidence and temporal robustness. 
Table~\ref{tab:scal-v} presents a rigorous evaluation of our proposed ECLIPS model in four transfer scenarios, following the same methodology as described in Table~\ref{tab:tab4}. The results demonstrate our model’s exemplary performance in terms of Bias and Variance. Notably, ECLIPS achieves the lowest average Bias and Varance, outperforming other methods, including several state-of-the-art (SOTA) models. These findings underscore the advanced capacity of ECLIPS to provide consistent and reliable predictions while effectively managing uncertainties, cementing its position as the new benchmark in the FAS domain.

\begin{table*}
\centering
\resizebox{\textwidth}{!}{% 
\begin{tabular}{lcccccccc|cc}
\Xhline{1pt}
\Gape[12pt][6pt]{\multirow{2}{*}{Method (\%)}} &
  \multicolumn{2}{c}{OCI$\rightarrow$M} &
  \multicolumn{2}{c}{OMI$\rightarrow$C} &
  \multicolumn{2}{c}{OCM$\rightarrow$I} &
  \multicolumn{2}{c|}{ICM$\rightarrow$O} &
  \multicolumn{2}{c}{Average} \\
                                  & HTER ($\downarrow$) & AUC ($\uparrow$) & HTER ($\downarrow$) & AUC ($\uparrow$)
                                  & HTER ($\downarrow$) & AUC ($\uparrow$) & HTER ($\downarrow$) & AUC ($\uparrow$) 
                                  & HTER ($\downarrow$) & AUC ($\uparrow$) \\ \Xhline{1pt}
ViT-B/16 \cite{dosovitskiy2020image}                        & 10.24 & 96.10 & 13.94	& 93.91 &  21.00 & 78.32 & 17.99 & 89.38 & 15.79 & 89.67 \\ 
ViT-B/16 (w/Sampling)                                       & 7.38 & 97.60 & 18.0 & 90.20 & 11.6 & 95.39 & 23.03 & 84.91 & 15.00 &  91.52 \\  
ViT-B/16 (w/Monte Carlo Dropout)                            & 7.14 & 98.19 & 14.66 & 93.53 & 12.5 & 93.88 & 25.46 & 82.47 & 14.19 & 92.76 \\ \Xhline{1pt}
CLIP-V \cite{radford2021learning}                           & 7.3 & 96.71 & 2.6 & 98.95 & 5.0 & 99.06 & 4.44 & 98.83 & 4.84 & 98.13 \\                   
CLIP-V (w/Sampling)                                         & 5.4 & 98.60 & 1.88 & 99.20 & 3.5 & 99.63 & 4.04 & 98.74 & 3.9 & 98.79 \\
FLIP-MCL (ICCV' 23) \cite{srivatsan2023flip}                & 4.95  & 98.11 & 0.54  & 99.98 & 4.25  & 99.07 & 2.31  & 99.63 & 3.01 & 99.19 \\ \Xhline{1pt} \rowcolor{green!10}
ECLIPS                                                      & 1.43  & 99.87 & 0.78  & 99.18 & 1.60  & 99.16 & 3.21  & 99.35 & 1.76 & 99.39 \\ \Xhline{1pt}
\end{tabular}}
\caption{
Ablation studies regarding model ensembling. Compared to the previous method applying a ViT-B/16 backbone, our proposed ECLIPS achieved significant quantitative improvement. Furthermore, ECLIPS achieved superior performance compared to CLIP CLIP-based method although we only apply Visual Encoder of the CLIP, different from the other methods.
}
\label{tab:tab-ensem}
\end{table*}
\subsection{Real-World Use-Case Performance}
The CelebA-Spoof dataset, constructed on the foundation of the CelebA dataset, is specifically designed for large-scale face recognition and anti-spoofing research, incorporating a spectrum of spoofing attacks. It encompasses a variety of imaging conditions, including lighting, pose, expression, and the presence or absence of makeup, making it a robust resource for evaluating the generalizability of models. SiW-Mv2, addresses face spoofing in real-world scenarios, incorporating various attack types and conditions indoors and outdoors. It plays a pivotal role in testing the resilience of deep learning-based anti-spoofing systems against the intricacies of real-world variability. Leveraging models trained on the OCIM protocol, testing on the public CelebA-Spoof dataset, which consists of individual frames sharing the same identity in a single video, allows for the validation of HTER and AUC performance metrics.

Referring to Table~\ref{tab:tab6}, we demonstrate enhanced performance over existing state-of-the-art models on this new dataset. SiW-Mv2's sequential frame composition within single videos enables a comprehensive assessment of not just HTER and AUC but also Variance and Bias. Models trained using the OCIM protocol and tested on the public SiW-Mv2 dataset report superior performance and generalization, surpassing that of current state-of-the-art models, thus advancing the field of FAS technology.

\subsection{Abulation Study}
As shown in Table~\ref{tab:tab-ensem}, we verify the ensemble effect by utilizing only the Visual Encoder from CLIP models, confirming its efficacy in Vision Transformer (ViT)implementations as well. The CLIP model, known for its superior feature representation capabilities, forms a part of our ensemble backbone. Through rigorous evaluation on both a new benchmark dataset and established datasets, our findings reveal that this ensemble strategy, when combined with the advanced feature representation of the CLIP backbone, markedly surpasses existing state-of-the-art methods. Our model not only excels in traditional performance metrics but also shines in our newly introduced generalization metrics, offering a more comprehensive and dependable assessment of performance across different testing conditions.
\subsection{Temporal robustness and Accuracy}
\begin{figure}[t]
\begin{center}
\includegraphics[width=0.93\linewidth]{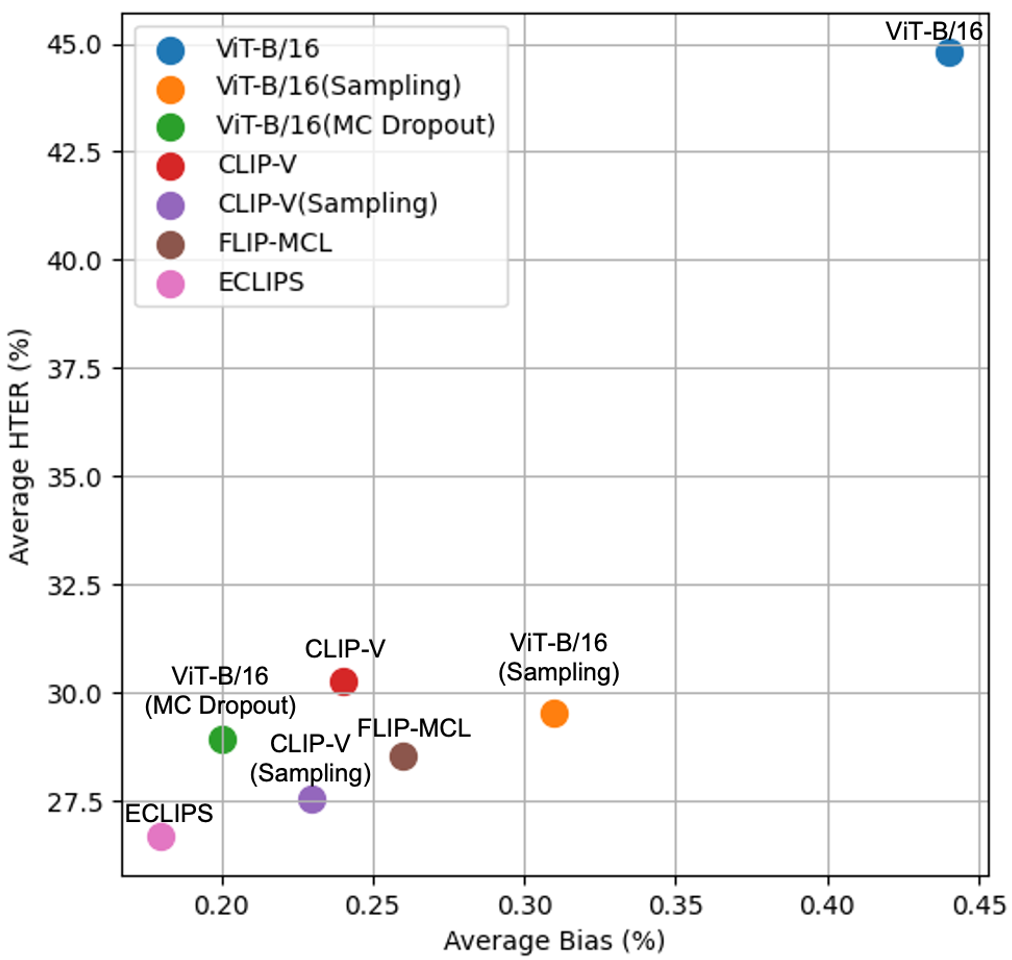}
\end{center}
\caption{The scatter plot of representative FAS models from Bias to HTER. 
Notably, the ECLIPS model is highlighted in the bottom-left, indicating its superior accuracy and generalization ability in FAS.} 
\label{fig:curve}
\end{figure}

Figure~\ref{fig:curve} illustrates the exceptional capabilities of the ECLIPS architecture in tackling overfitting, enhancing predictive performance, and demonstrating remarkable robustness against temporal and noise variations within FAS tasks. Notably, in the scatter plot comparing ensemble methods evaluated on OCIM→S, the ECLIPS model is positioned in the bottom-left quadrant. This positioning not only highlights its unparalleled accuracy but also its superior performance in terms of variance, showcasing its outstanding generalization capabilities and robustness in both accuracy and variance within the FAS domain. However, the results depicted in the figure also caution that a favorable bias does not necessarily correlate with good accuracy. Therefore, it’s imperative to consider temporal robustness metrics alongside accuracy for a comprehensive evaluation.

\section{Conclusions}
\label{sec:conclusion}
In this paper, we conducted a thorough analysis of the FAS model'scapabilities in detecting presentation attacks, taking into account dataset characteristics, evaluation metrics, and design principles.
Our investigation revealed significant limitations within the existing evaluation framework, especially regarding the assessment of generalization and model scalability. To verify and address these challenges, we utilized a wide range of FAS evaluation datasets reflective of real-world scenarios and introduced a robustness-aware evaluation metric designed to more accurately gauge the FAS model's generalization capabilities. Based on our analyses, we revisited and refined the ensemble strategy for smaller-sized FAS models, achieving state-of-the-art performance through extensive experimentation. Future research could explore the feasibility of analyzing spoofing scenes across diverse scenarios to enhance datasets in ways that bolster the effectiveness of the proposed ensemble method.

{
    \small
    \bibliographystyle{ieeenat_fullname}
    \bibliography{main}
}


\end{document}